\documentclass{article}

\usepackage[main, final]{neurips_2026}
\usepackage{multirow}
\usepackage{subcaption}

\usepackage[utf8]{inputenc} %
\usepackage[T1]{fontenc}    %
\usepackage{hyperref}       %
\usepackage{url}            %
\usepackage{booktabs}       %
\usepackage{amsfonts}       %
\usepackage{nicefrac}       %
\usepackage{microtype}      %
\usepackage{xcolor}         %
\usepackage{colortbl}
\definecolor{ours}{gray}{0.92}
\usepackage{graphicx}
\usepackage{amsmath} %
\usepackage{enumitem}
\usepackage{hyperref}
\usepackage[nameinlink,capitalize]{cleveref}
\usepackage{makecell}
\usepackage{caption}
\usepackage{placeins}
\usepackage{needspace}
\usepackage{wrapfig}
\newcommand{\new}[1]{#1}
\title{
Dense Latent Reward Alignment for \\Text-to-Image Flow Matching Models
}

\author{%
  Zeeshan Khan \qquad Xin Yu \qquad Shizhe Chen \qquad Cordelia Schmid\\
  Inria, \'Ecole normale sup\'erieure, CNRS, PSL Research University\\
  \href{https://zeeshank95.github.io/deflowcritic/}{\texttt{https://zeeshank95.github.io/deflowcritic}}
}

\begin{document}

\maketitle

\begin{abstract}
Online reinforcement learning (RL) has emerged as a powerful paradigm for aligning text-to-image flow matching models with complex user intent. 
However, existing methods typically rely on sparse rewards computed only from final decoded images. This delays credit assignment across the denoising trajectory and makes training computationally expensive. Moreover, simply combining multiple rewards does not reliably produce a Pareto trade-off, where gains in compositional accuracy often degrade aesthetic quality.
In this work, we propose \textbf{DeFlowCritic}, an efficient online RL framework based on Dense Latent Reward Alignment. Unlike prior work, our method provides dense supervision on intermediate latents, enabling early-stage guidance for the denoising trajectory. 
To achieve this, we introduce an internal critic that evaluates noisy latents using features extracted directly from the diffusion model, a representation inherently better suited for latent-space reward estimation than standard image encoders.
We construct a high-quality reward-modeling dataset with continuous scores for critic training, \new{capturing compositional correctness on high-quality reference images}.
The resulting dense reward enables efficient online RL training and inference-time scaling. 
Experiments show that our method achieves a superior balance between prompt faithfulness and visual quality on GenEval2 and T2I-CompBench++ benchmarks. 
The approach also generalizes across text-to-image flow matching models, including SD3.5-Medium, SD3.5-Large, and FLUX.1-dev.

\end{abstract}

\section{Introduction}
\label{sec:intro}

While state-of-the-art text-to-image (T2I) models based on diffusion and flow matching~\citep{esser2024sd35, lipman2022flowmatching,liu2022rectifiedflow} 
achieve impressive photorealism,  they consistently struggle with  \emph{compositional prompts} that require precise numeracy, attribute binding, and spatial coordination.
Closing this gap is essential for moving beyond visually appealing generation toward reliable, intent-aligned image synthesis.

Online reinforcement learning (RL) has recently become one of the most effective approaches for improving compositional alignment in text-to-image generation. 
Methods such as DDPO~\citep{black2023ddpo}, DPOK~\citep{fan2024dpok}, and Flow-GRPO~\citep{liu2025flowgrpo} directly optimize generative models using rewards computed from final decoded images, yielding strong gains on compositional benchmarks such as GenEval~\citep{ghosh2023geneval} and T2I-CompBench~\citep{huang2023t2i}.

However, existing online RL methods remain expensive and suffer from sparse credit assignment. 
Each rollout must complete an entire denoising trajectory before receiving a terminal reward, and group-relative methods such as GRPO~\citep{shao2024deepseekmath} further multiply this cost by sampling multiple trajectories per prompt. 
    Efficiency-oriented variants~\citep{he2025tempflowgrpo, Yan_2025_ICCV, gdpo2025} reduce overhead by simplifying the training procedure, for example by removing classifier-free guidance~\citep{ho2022classifier} or KL regularization~\citep{relative2010}, but these changes can degrade image quality and sample diversity. 
Moreover, a single terminal reward provides no direct indication of which denoising steps caused a failure. DenseGRPO~\citep{densegrpo2026} addresses this credit-assignment issue with step-level rewards, but its reward construction remains costly: each intermediate latent is further denoised into a clean image before being evaluated by an image-space reward model.

Designing the reward itself is equally challenging. Optimizing only for compositional accuracy can produce images that satisfy the prompt while sacrificing physical plausibility or visual appeal. 
A natural remedy is to combine a compositional reward with human-preference models such as PickScore~\citep{kirstain2023pickapic} or HPSv2~\citep{wu2023hpsv2}; however, these rewards may differ in scale and optimization direction, making clean Pareto trade-offs difficult. 
GDPO~\citep{gdpo2025} mitigates this issue through reward-specific normalization, but still requires careful reward-weight tuning and relies on sparse terminal feedback.
Recent work has explored latent reward models that score noisy denoising states directly~\citep{slrm2025consistent, zhang2025latentreward}. For example, SLRM~\citep{slrm2025consistent} trains a latent reward model from human-preference data such as Pick-a-Pic~\citep{kirstain2023pickapic}. While such supervision is useful for aesthetic or preference-based quality, it does not explicitly encode compositional correctness. Learning a dense latent reward that jointly captures prompt faithfulness and visual quality therefore remains an open problem.

In this work, we propose \textbf{DeFlowCritic}, a latent critic for flow matching text-to-image models that improves compositional alignment while retaining visual quality. 
We first construct a new reward-modeling dataset, where each training example consists of a high-quality reference image, one correct prompt, and several systematically perturbed negative prompts that introduce controlled compositional errors. 
The critic model is based on the flow matching model itself, which contains stronger representations between noisy latents and texts to score noisy intermediate states. 
We train it with a soft classification loss to calibrate absolute reward scores, a graded ranking loss to enforce the ordering between correct and corrupted prompts, and a positive-only flow-matching auxiliary to keep the adapted backbone close to the diffusion manifold. 
Together, this dataset and objective yield a single latent reward that captures prompt faithfulness and favors images close to the distribution of the high-quality reference images.

We apply the resulting critic in two settings. 
First, we use it as a dense per-step reward in online GRPO training for flow-matching T2I models. This improves credit assignment, especially at early denoising steps, while avoiding the need to decode intermediate latents into clean images. 
Second, we use the same critic for inference-time scaling, where multiple latent trajectories are explored and selected according to their latent rewards, again without repeated image decoding. 
Experiments on GenEval2 and T2I-CompBench++ benchmarks show that DeFlowCritic improves the balance of compositional accuracy, visual quality, and efficiency in both online RL and inference-time scaling. 
We demonstrate its applicability to different T2I backbones, including SD3.5-Medium, SD3.5-Large and FLUX.1-dev.

Our contributions are as follows:
\begin{itemize}[itemsep=0.1em,parsep=0em,topsep=0em,partopsep=0em]
    \item We introduce DeFlowCritic, a dense latent reward model trained on a high-quality reward-modeling dataset to score intermediate denoising latents.
    
    \item We leverage the critic model to improve two orthogonal settings: as a dense reward for online GRPO fine-tuning and as a selection signal for inference-time scaling.

    \item We demonstrate a better balance of compositional alignment, visual quality, and efficiency than state-of-the-art methods across compositional T2I benchmarks, and validate our approach across multiple T2I backbones.
    
\end{itemize}

\section{Related Work}
\label{sec:related}

\noindent\textbf{Compositional text-to-image generation.}
Current text-to-image models based on diffusion and flow matching \citep{song2020sde,lipman2022flowmatching,liu2022rectifiedflow,esser2024sd35} can synthesize highly realistic images but remain unreliable on complex, out-of-distribution compositional prompts~\citep{ghosh2023geneval,kamath2025geneval2,huang2023t2i,huang2025compbench}. 
Training-free methods improve compositionality at inference time through layout guidance~\citep{li2023gligen}, attention manipulation~\citep{chefer2023attend}, or inference-time scaling~\citep{kim2025inferencetimescaling,singhal2024fksteering,ma2025inferencetimediffusion} which allocate additional test-time compute to explore and select better generations. In contrast, training-based methods directly adapt the model using supervised fine-tuning~\citep{ouyang2022training, chung2022finetuned, wei2022finetuned}, preference optimization~\citep{rafailov2023dpo, fan2024dpok, wallace2024diffusiondpo}, or reinforcement learning (RL)~\citep{densegrpo2026,liu2025flowgrpo, he2025tempflowgrpo}, but typically rely on rewards computed from final decoded images. 

\noindent\textbf{Alignment for text-to-image flow-matching models.}
In text-to-image generation, existing alignment methods can be broadly grouped into reward backpropagation, offline preference optimization, and online reinforcement learning. 
Reward backpropagation methods~\citep{prabhudesai2023rewardbp,clark2023differentiablereward} directly optimize differentiable reward signals, but require gradients through the sampler. 
Offline preference methods, including Diffusion-DPO~\citep{wallace2024diffusiondpo} and related DPO-style objectives~\citep{rafailov2023dpo,yang2024d3po,liang2025spoaesthetic}, avoid online rollouts but are constrained by the coverage and quality of static preference data. 
Online policy-gradient methods such as DDPO~\citep{black2023ddpo}, DPOK~\citep{fan2024dpok}, and Flow-GRPO~\citep{liu2025flowgrpo} instead cast denoising as a sequential decision process and achieve strong alignment gains, but incur high computational cost and suffer from sparse rewards. 
Recent methods address these limitations through fewer sampling steps, modified stochastic windows, or improved advantage estimation~\citep{Yan_2025_ICCV,gdpo2025}. DenseGRPO~\citep{densegrpo2026} further tackles sparse credit assignment with step-level rewards, but is more expensive by transforming intermediate states into clean images for reward calculation. 
In contrast, DeFlowCritic targets this reward bottleneck by learning a critic that directly scores noisy latents, enabling efficient dense reward computation.

\noindent\textbf{Reward modeling.}
Most reward models for image generation operate on final decoded images. 
Human-preference models such as PickScore~\citep{kirstain2023pickapic}, HPSv2/v3~\citep{wu2023hpsv2,ma2025hpsv3}, ImageReward~\citep{xu2024imagereward}, MPS~\citep{zhang2024mps}, VisionReward~\citep{xu2024visionreward}, and MLLM-based unified reward models~\citep{wang2025unifiedreward} provide useful terminal feedback, but are typically trained on large-scale image-level comparisons and are costly to query during online RL. 
More importantly, terminal rewards do not indicate which denoising steps caused a failure. 
Step-level preference methods\citep{liang2025spoaesthetic,zhang2024onlinevpo} densify supervision, but perform scoring in decoded pixel space with image encoders, which is less efficient and less reliable for noisy intermediate states.
Recent latent reward models, including SLRM~\citep{slrm2025consistent} and noise-aware latent reward modeling~\citep{zhang2025latentreward}, show that a diffusion backbone scores noisy latents more reliably than an external image encoder. They are trained on large-scale human preference pairs, and therefore mainly capture overall quality preference rather than fine-grained prompt faithfulness. We build on the SLRM architecture and adapt it to compositional alignment. 
We automatically construct a dataset of 5K reference images with lightweight human verification and improve the training objective to learn a more effective compositional critic.

\section{Method}
\label{sec:method}

We train a latent reward model DeFlowCritic that provides dense alignment signals for the denoising trajectory of flow-matching text-to-image models, balancing compositional fidelity and visual quality. 
We first describe the curated reward-modeling dataset used for training and evaluation (\Cref{sec:reward-data}). 
We then present the proposed  model for predicting calibrated latent rewards (\Cref{sec:critic_model}). 
Finally, we show how the critic supports text-to-image generation in two settings: online Flow-GRPO training with  per-step rewards and reward-guided search at inference time (\Cref{sec:stage2}).

\begin{figure*}[t]
  \centering
  \captionsetup[figure]{belowskip=0pt}
  \includegraphics[width=\textwidth]{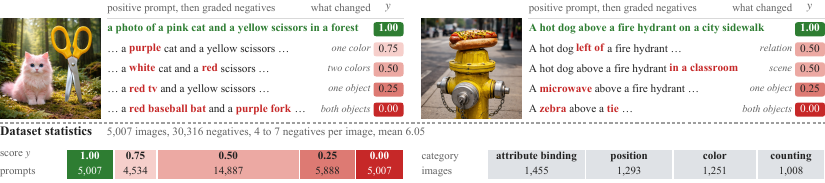}
 \caption{Samples from our \textsc{GPT-Image-Composition} dataset and its training-set statistics. Four negatives are shown per image, with corrupted words in red and ``\dots'' marking unchanged words.}
  \label{fig:dataset}
  \vspace{-12pt}
\end{figure*}

\subsection{Reward Modeling Dataset}
\label{sec:reward-data}

Existing text-to-image reward datasets, such as Pick-a-Pic~\citep{kirstain2023pickapic}, primarily capture image-level human preference with positive and negative images, rather than fine-grained compositional faithfulness. We therefore construct \textsc{GPT-Image-Composition}, a reward-modeling dataset that captures both visual quality and prompt-level compositional correctness.

\noindent\textbf{Positive image-text pairs.}
We first generate prompts from compositional templates spanning four categories: counting, color, spatial position, and attribute binding, following the compositional structure of GenEval~\citep{ghosh2023geneval}. 
For each prompt $c^+$, we use GPT-Image-1.5 to generate a high-quality reference image $I$. 
We manually check the generated data, removing approximately $25-30\%$ of the images that contain incorrect compositions, visible artifacts, or undesired styles such as poor lighting, overly dark scenes, or cartoon-like rendering.
The positive pair $(c^+, I)$ is assigned the ground-truth score $y=1$.

\noindent\textbf{Negative image-text pairs.}
Based on the positive image-text pair $(c^+, I)$, we construct \new{up to seven} negative prompts by systematically perturbing one or more attribute slots in $c^+$, including object identity, counts, attributes, and spatial relations. 
We assign each negative pair $(c^-_*, I)$ a continuous compositional score $y \in [0, 1)$ measuring agreement with the fixed reference image. Lower scores reflect increasingly severe compositional mismatches. 
For example, for $c^+=$ ``a photo of a pink cat and a yellow scissors in a forest'' (\Cref{fig:dataset}, left), changing one color (``a purple cat'') gives $y=0.75$, changing both colors (``a white cat and a red scissors'') gives $y=0.5$, replacing one object and its color (``a red tv'') gives $y=0.25$, and replacing both objects gives $y=0$.

\noindent\textbf{Dataset summary.}
As shown in Fig.~\ref{fig:dataset}, the resulting dataset contains $5$K visually clean reference images, each paired with one positive prompt and $4$ to $7$ controlled negative prompts. 
These negative prompts introduce different levels of compositional violations, enabling the critic to learn fine-grained distinctions between well-aligned and partially mismatched image-prompt pairs. 
We split the dataset into training and test sets with  $5{,}007$ and $240$ examples, respectively. The evaluation split is balanced across the query categories (counting, color, attribute binding, and spatial position).

\begin{figure}[t]
    \centering
    \includegraphics[width=\textwidth]{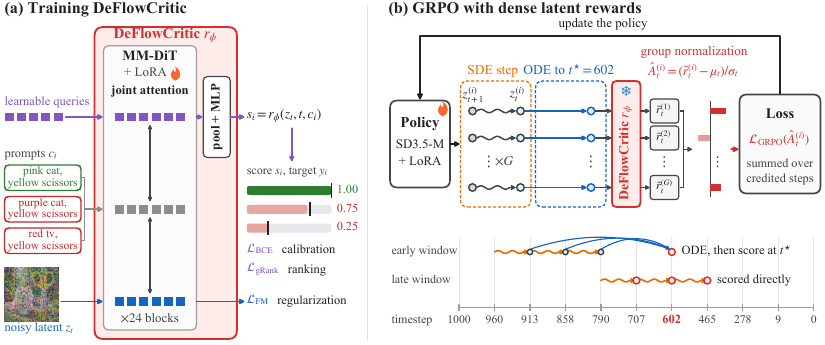}
    \caption{(a) DeFlowCritic scores a noisy latent $z_t$ against a positive prompt and several negatives. Each latent-prompt pair $(z_t, c_i)$ is fed to the critic as a separate input. Bars show $s_i$ (predicted) and ticks show $y_i$ (target). $\mathcal{L}_{\mathrm{FM}}$ is applied to the positive prompt only. (b) In GRPO, the frozen critic provides dense rewards by scoring the latent at every SDE step, after an ODE rollout for early steps.}
    \label{fig:method}
\end{figure}

\subsection{DeFlowCritic: Dense Latent Reward Model}
\label{sec:critic_model}

Let $z_t$ denote the noisy latent at denoising timestep $t$, and $c$ the text prompt. The latent reward model $r_\phi$ predicts a scalar reward $s_i = r_\phi(z_t,t,c_i)$ that measures the quality of the latent-prompt pair.

\subsubsection{Model Architecture}
We build the reward model on top of the SD3.5-Medium MM-DiT backbone~\citep{esser2024sd35} following SLRM~\citep{slrm2025consistent}, which jointly represents noisy image latents and text tokens and therefore provides features well suited for latent-space alignment. 
We augment the frozen backbone with two trainable components: (i) LoRA~\citep{hu2022lora} adapters applied to the attention projection and feed-forward layers in each MM-DiT block, 
and (ii) $N_q$ learnable query tokens concatenated with the text stream inside the MM-DiT joint-attention blocks, where they attend to both text and image-latent tokens to summarize cross-modal alignment information.

After the final transformer block, the query tokens are mean-pooled and passed through a lightweight MLP head with sigmoid activation:
\begin{equation}
    r_\phi(z_t, t, c) = \sigma\!\left(\mathrm{MLP}\!\left(
    \frac{1}{N_q}\sum_{k=1}^{N_q} q_k^{(L)}(z_t, t, c)
    \right)\right),
    \label{eq:reward}
\end{equation}
where $q_k^{(L)}$ denotes the $k$-th query token after the last block.

\subsubsection{Training Objective}
\label{sec:reward-loss}

Let $s_i$ denote the predicted reward and let $y_i\in [0,1]$ denote the continuous score defined in \Cref{sec:reward-data}. 
We train the critic with a composite objective on the constructed dataset:
\begin{equation}
    \mathcal{L}_{r_\phi}
    =
    \mathcal{L}_{\mathrm{BCE}}
    + \lambda_{\mathrm{rank}}\mathcal{L}_{\mathrm{gRank}}
    + \lambda_{\mathrm{FM}}\mathcal{L}_{\mathrm{FM}}.
    \label{eq:reward-loss}
\end{equation}
The three terms play complementary roles: score calibration fixes the absolute reward scale, compositional ranking enforces the ordering among prompts paired with the same image, and flow-matching regularization keeps the adapted backbone aligned with the denoising manifold. Figure~\ref{fig:method}(a) shows how DeFlowCritic is trained with the three objectives.

\noindent\textbf{Score prediction loss $\mathcal{L}_{\mathrm{BCE}}$.}
We use soft binary cross-entropy on the continuous targets:
\begin{equation}
    \mathcal{L}_{\mathrm{BCE}}
    =
    -\mathbb{E}_i
    \left[
        y_i\log s_i + (1-y_i)\log(1-s_i)
    \right].
    \label{eq:soft-bce}
\end{equation}
This term allows the critic to assign more calibrated scores to latent-prompt pairs with different levels of compositional agreement.

\noindent\textbf{Ranking loss $\mathcal{L}_{\mathrm{gRank}}$.}
For positive and negative prompts paired with the same reference image, relative ordering provides strong supervision to measure the quality. 
We therefore apply a graded RankNet-style pairwise loss~\citep{burges2005ranknet}:
\begin{equation}
    \mathcal{L}_{\mathrm{gRank}}
    =
    \mathbb{E}_{\mathrm{rows}}
    \frac{
    \sum_{(i,j):\,y_i>y_j}
    (y_i-y_j)\,
    \mathrm{softplus}\!\left(-\frac{s_i-s_j}{\tau}\right)}
    {
    \sum_{(i,j):\,y_i>y_j} (y_i-y_j)
    },
    \qquad \tau=0.15 .
    \label{eq:grank}
\end{equation}

\noindent\textbf{Flow-matching regularization $\mathcal{L}_{\mathrm{FM}}$.}
Since the reward model is trained on a small high-quality dataset, preference losses alone may shift the adapted backbone too far from the denoising features. We therefore add a flow-matching auxiliary loss on positive prompts as regularization:
\begin{equation}
    \mathcal{L}_{\mathrm{FM}}
    =
    \mathbb{E}_{t,z_t,c^+}
    \left[
        w(t)
        \left\|
        v_\phi(z_t,t,c^+) - u_t(z_t\mid z_0)
        \right\|_2^2
    \right],
    \qquad
    w(t)=1+\alpha_{\mathrm{ramp}}
    \frac{t-t_{\min}}{t_{\max}-t_{\min}} .
    \label{eq:fm-anchor}
\end{equation}
Here $u_t$ is the target conditional velocity for the clean reference latent $z_0$, $\alpha_{\mathrm{ramp}}=2.0$, and timesteps are sampled from $[t_{\min},t_{\max}]=[50,950]$ to avoid the extreme ends of the noise schedule. 
The ramp up-weights noisier timesteps, where velocity prediction provides a stronger constraint on the latent representation.
Applying this auxiliary only to the positive prompt $c^+$ avoids training the model to denoise a clean image under an intentionally mismatched text condition.

\subsection{Improving T2I Generation with DeFlowCritic}
\label{sec:stage2}

\paragraph{Dense latent rewards for Flow-GRPO.}  
We use the same rollout and policy-update machinery as
Flow-GRPO~\citep{liu2025flowgrpo}: for each prompt, the policy samples a
group of stochastic trajectories within the SDE window, recomputes
per-step log-probabilities, and optimizes the clipped GRPO surrogate over
the credited denoising steps. 

Our modification is restricted to advantage construction. Standard Flow-GRPO assigns each trajectory a single terminal reward, normalizes these rewards across the group, and broadcasts the same advantage to every credited timestep. In contrast, we query the frozen latent critic at each credited timestep and normalize rewards across the group separately for each $t$, yielding timestep-specific advantages, as illustrated in Fig.~\ref{fig:method}(b).

Because the critic is best calibrated near a trusted low-noise timestep
$t^\star$ (we use $t^\star=602$), we do not score high-noise latents
directly. \new{Steps that land on $t\in\{707,602,465\}$ are close to this range and are scored directly.} Each \new{earlier} credited latent $z_t^{(i)}$ is first mapped to
$t^\star$ by a short deterministic Euler rollout and then evaluated by the
critic and we compute the group-relative advantage for each timestep $t$:
\begin{equation}
    \tilde r_t^{(i)}
    =
    r_\phi\!\left(\mathrm{ODE}_{t\rightarrow t^\star}(z_t^{(i)}),\, t^\star,\, c\right),
    \quad
    \hat A_t^{(i)}
    =
    \frac{\tilde r_t^{(i)}-\mu_t}{\sigma_t+\epsilon_{\mathrm{std}}},
    \quad
    \mu_t,\sigma_t
    =
    \mathrm{mean},\mathrm{std}_{i=1}^{G}
    \left(\tilde r_t^{(i)}\right).
    \label{eq:ode-reward}
\end{equation}
\new{We} plug $\hat A_t^{(i)}$ into the unchanged Flow-GRPO surrogate. 
This replaces a terminal image-level reward broadcast with step-specific
latent feedback, while avoiding VAE decoding and external pixel-space
reward-model calls inside the dense reward loop.

\paragraph{Guidance for inference-time scaling.}
We use the frozen critic for test-time trajectory selection: instead of sampling a single denoising trajectory, we denoise $K$ independent candidates only up to the scoring timestep $t^\star$, rank them with $r_\phi$, and complete only the best candidate with deterministic ODE steps.
Losing candidates are thus pruned before being fully denoised or decoded, whereas an image-space reward model must finish and decode all $K$ samples before it can rank them.

\section{Experiments}
\label{sec:experiments}
\begin{table*}[t]
\centering
\small
\setlength{\tabcolsep}{3pt}
\renewcommand{\arraystretch}{1.15}
\caption{
Comparison of Flow-GRPO variants on GenEval2 and T2I-CompBench++ benchmarks. 
All models are built upon the SD3.5-M~\cite{esser2024sd35} baseline. We report compositional fidelity, image quality measured by LAION Aesthetic~\citep{schuhmann2022laion} and HPSv3~\citep{ma2025hpsv3}, and relative training time.
}
\label{tab:main_comparison}
\begin{tabular}{lccccccc}
\toprule
\multirow{2}{*}{Models}
& \multicolumn{3}{c}{GenEval2}
& \multicolumn{3}{c}{T2I-CompBench++}
& \multirow{2}{*}{Time $\downarrow$} \\
\cmidrule(lr){2-4} \cmidrule(lr){5-7}
 & Comp.\,$\uparrow$ & Aesth.\,$\uparrow$ & HPSv3\,$\uparrow$
 & Comp.\,$\uparrow$ & Aesth.\,$\uparrow$ & HPSv3\,$\uparrow$
 &  \\
\midrule
SD3.5-M ~\cite{esser2024sd35}                           & 18.10 & 5.51 & 6.66  & 52.37 & 5.46 & 7.76  & --  \\
Flow-GRPO (GenEval) ~\cite{liu2025flowgrpo}               & 23.40 & 5.41 & 7.13  & 59.44 & 5.27 & 9.26  & 1$\times$  \\
fast Flow-GRPO (GenEval)           & 22.70 & 5.17 & 4.45  & 58.64 & 5.18 & 5.71  & 0.1$\times$ \\
fast Flow-GRPO (GenEval + Aesth.)  & 12.50 & 6.91 & 6.89  & 55.24 & 6.96 & 8.72  & 0.12$\times$ \\
\rowcolor{ours} fast Flow-GRPO (DeFlowCritic) & 21.44 & 5.66 & 8.07 & 59.37 & 5.69 & 10.34 & 0.1$\times$ \\
\bottomrule
\end{tabular}
\vspace{-6pt}
\end{table*}

\subsection{Implementation Details}
\label{sec:impl}

\paragraph{DeFlowCritic training.}
We train DeFlowCritic on our reward-modeling dataset with a frozen
SD3.5-Medium MM-DiT backbone. The trainable parameters are LoRA adapters,
a reward head, and $N_q=16$ query tokens injected into every transformer
block. We use LoRA rank $32$ with $\alpha=64$, train for \new{up to} $20$ epochs, and
sample timesteps from $[50,950]$. The reward head and LoRA adapters use
learning rates $1{\times}10^{-4}$ and $3{\times}10^{-5}$, respectively. 
We set $\lambda_{\mathrm{rank}}=0.2$ and $\lambda_{\mathrm{FM}}=1$ in \Cref{eq:reward-loss}. 
We stop training once ranking accuracy on a held-out split plateaus, which takes around $8$ H100-hours (\Cref{app:critic-training}).

\paragraph{GRPO training.}
We fine-tune SD3.5-Medium with Flow-GRPO using DeFlowCritic as the dense
latent reward. Training uses GenEval prompts, group size $G=22$, rollout
length $T=10$, and resolution $512{\times}512$; evaluation uses
$T_{\mathrm{eval}}=40$ denoising steps. Dense rewards are computed with
ODE-lookahead in \Cref{eq:ode-reward}, using trusted timestep
$t^\star=602$ and scoring steps $\{465, 602, 707\}$. We disable classifier-free guidance and KL regularization during training, and use
AdamW with learning rate $3{\times}10^{-4}$. GRPO training takes approx. 90 H100 hours.

\paragraph{Evaluation.}
Our policy is trained on GenEval prompts, so GenEval-1 is in-domain and close to saturated. We use it only to compare with the published results from other latent reward models (\Cref{tab:latent_rewards}), and report our main results zero-shot on GenEval2~\citep{kamath2025geneval2} and T2I-CompBench++~\citep{huang2025compbench}, which neither the critic nor the policy sees during training. For compositional fidelity, we report the aggregate compositional score from each benchmark. For image quality, we
report LAION Aesthetic Score~\citep{schuhmann2022laion} and
HPSv3~\citep{ma2025hpsv3}.

\subsection{Comparison with State-of-the-Art Methods}
\label{sec:sota}

\paragraph{Baselines.}
We compare SD3.5-Medium variants that share the same backbone but differ in
post-training reward and regularization. \emph{SD3.5-M}~\citep{esser2024sd35}
is the pretrained model. \emph{Flow-GRPO (GenEval)}~\citep{liu2025flowgrpo}
uses terminal GenEval rewards with CFG and KL regularization. \emph{Fast
Flow-GRPO (GenEval)} removes CFG and KL for faster training. \emph{Fast
Flow-GRPO (GenEval + Aesth.)}  adds a terminal LAION aesthetic reward to the
fast setting. Our method uses the same fast GRPO setting but replaces
terminal rewards with dense latent rewards from DeFlowCritic.

\paragraph{Comparison with GRPO methods.}
\Cref{tab:main_comparison} reports results on GenEval2 and
T2I-CompBench++. DeFlowCritic
matches standard Flow-GRPO on T2I-CompBench++ compositional fidelity
($59.37$ vs.\ $59.44$) and remains competitive on GenEval2 ($21.44$ vs.\
$23.40$), while improving image-quality metrics on both benchmarks. Compared
with Flow-GRPO, DeFlowCritic improves HPSv3 from $7.13$ to $8.07$ on
GenEval2 and from $9.26$ to $10.34$ on T2I-CompBench++. Compared with Fast
Flow-GRPO, the gains are larger, indicating that dense latent rewards help
stabilize fast training when CFG and KL are removed. The aesthetic-reward
baseline improves LAION Aesthetic but significantly  reduces GenEval2 compositional
accuracy, suggesting that naive terminal multi-reward mixing does not yield a
clean quality-composition trade-off. DeFlowCritic obtains a stronger
trade-off while achieving a $10\times$ training speed-up over standard
Flow-GRPO at matched group size and inference schedule.

\begin{table*}[t]
\centering
\caption{Category-level comparison on T2I-CompBench++~\citep{huang2025compbench}. Compositional fidelity is reported.}
\label{tab:t2i-compbench}
\footnotesize
\setlength{\tabcolsep}{4pt}
\renewcommand{\arraystretch}{1.12}
\resizebox{\textwidth}{!}{%
\begin{tabular}{lccccccc}
\toprule
\multirow{2}{*}{Models}
& \multicolumn{7}{c}{T2I-CompBench++ Categories} \\
\cmidrule(lr){2-8}
& Color
& Shape
& Texture
& 2D-Spatial
& 3D-Spatial
& Numeracy
& Non-Spatial \\
\midrule
SD3.5-M ~\cite{esser2024sd35}                          & 79.94 & 56.69 & 73.38 & 28.50 & 37.39 & 59.27 & 31.46 \\
Flow-GRPO (GenEval)  ~\cite{liu2025flowgrpo}              & 83.79 & 61.30 & 72.36 & 54.47 & 44.71 & 67.52 & 31.95 \\
fast Flow-GRPO (GenEval)           & 82.70 & 64.90 & 72.60 & 51.10 & 41.00 & 67.10 & 31.00 \\
fast Flow-GRPO (GenEval + Aesth.)  & 81.50 & 57.40 & 67.40 & 42.10 & 43.90 & 64.10 & 30.30 \\
	\rowcolor{ours} fast Flow-GRPO (DeFlowCritic) & 85.38 & 62.66 & 74.13 & 50.27 & 45.38 & 66.46 & 31.37 \\
\bottomrule
\end{tabular}%
\vspace{-12pt}
}
\end{table*}

\begin{figure*}[t]
  \centering
  {\scriptsize
\setlength{\tabcolsep}{1.5pt}
\renewcommand{\arraystretch}{0.5}
\def\qi#1{\raisebox{-.5\height}{\includegraphics[width=0.205\textwidth]{camera_ready/figures/qual_sd35/#1}}}
\def\qp#1{\parbox{0.12\textwidth}{\centering\emph{#1}}}
\def\qh#1{\parbox[b]{0.205\textwidth}{\centering #1}}
\begin{tabular}{@{}c@{\hspace{3pt}}cccc@{}}
 & \qh{Flow-GRPO\\(GenEval)} & \qh{fast Flow-GRPO\\(GenEval)} & \qh{fast Flow-GRPO\\(GenEval + Aesth.)} & \qh{Ours} \\ \noalign{\vskip 3pt}
\qp{a photo of three persons} & \qi{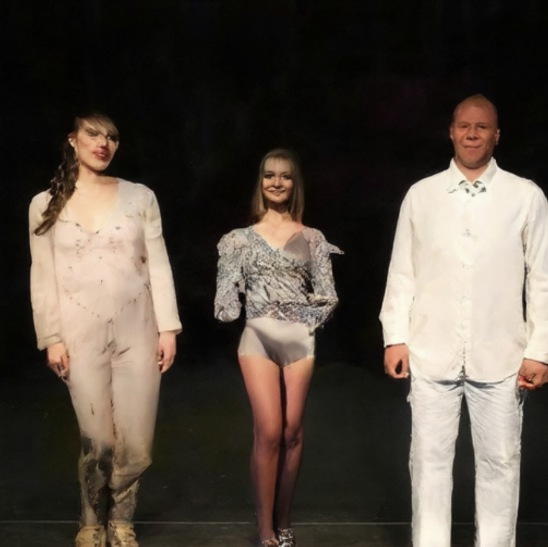} & \qi{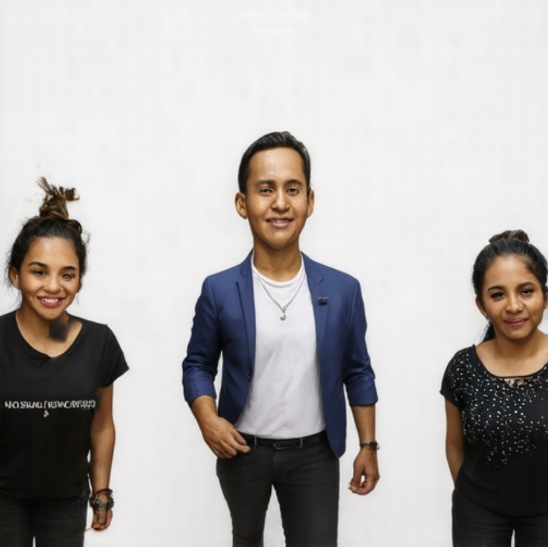} & \qi{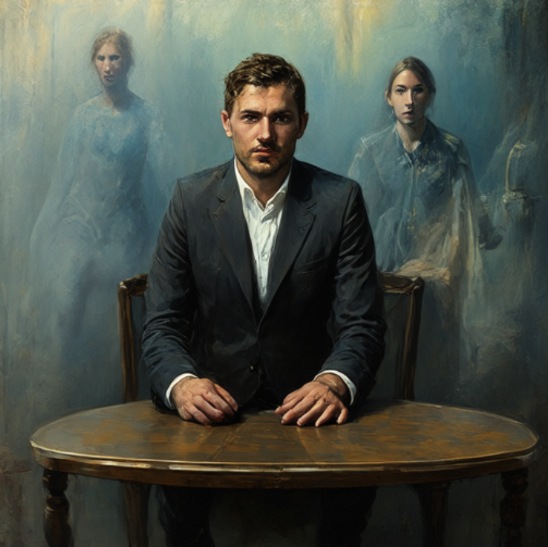} & \qi{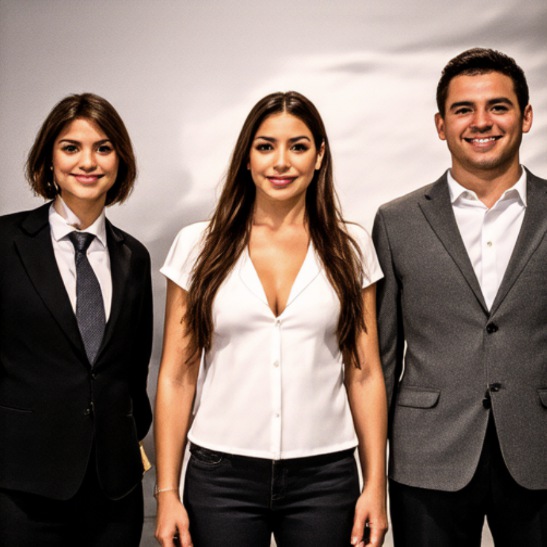} \\ \noalign{\vskip 2pt}
\qp{a photo of a couch below a vase} & \qi{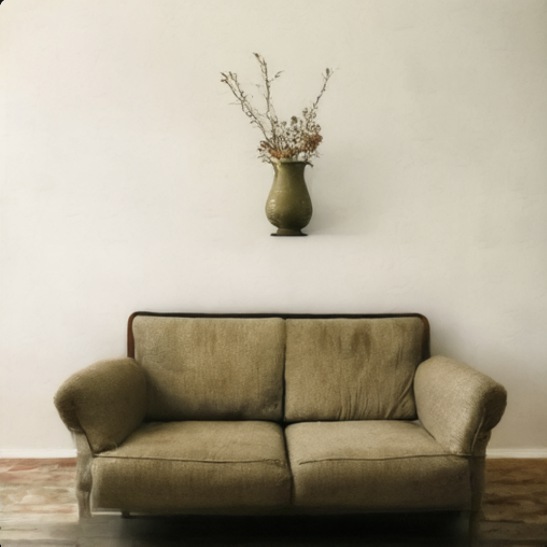} & \qi{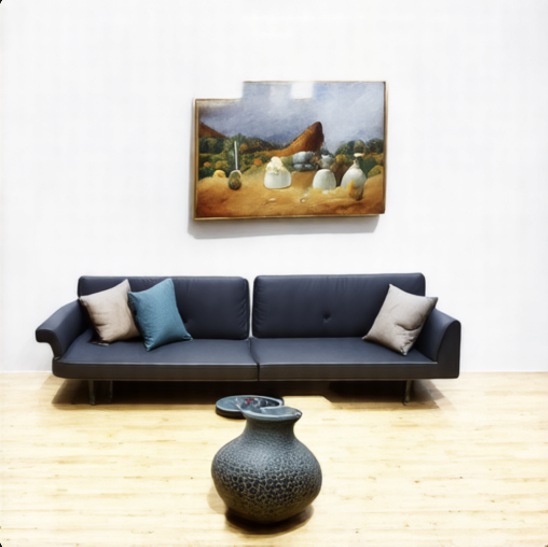} & \qi{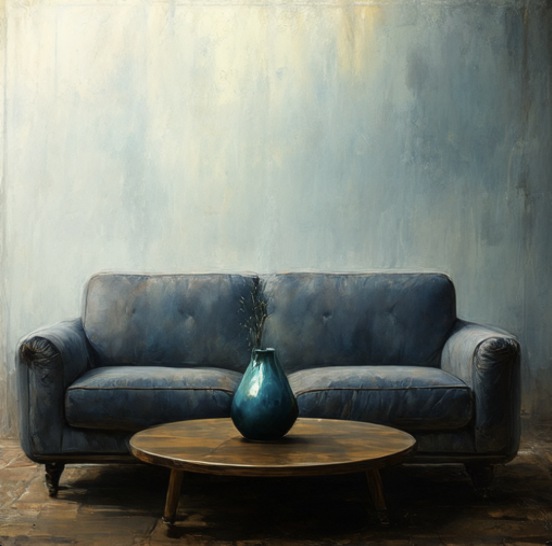} & \qi{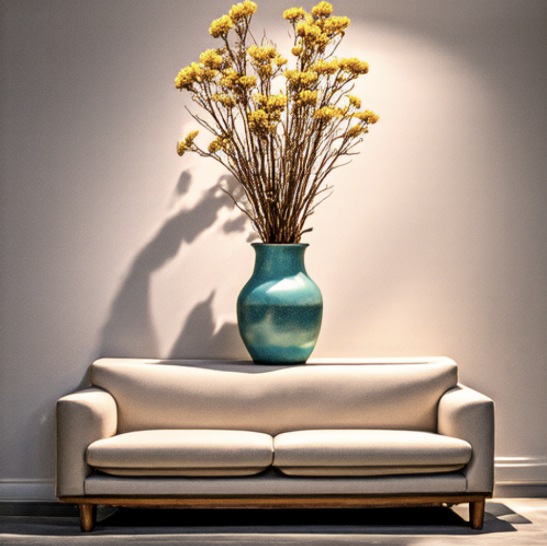} \\ \noalign{\vskip 2pt}
\qp{a brown turtle on top of a horse} & \qi{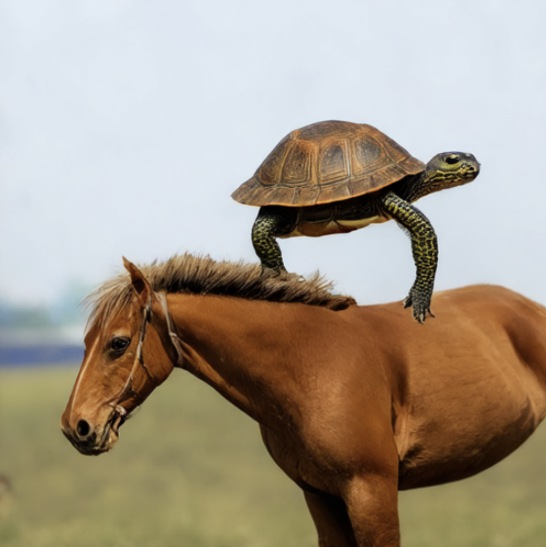} & \qi{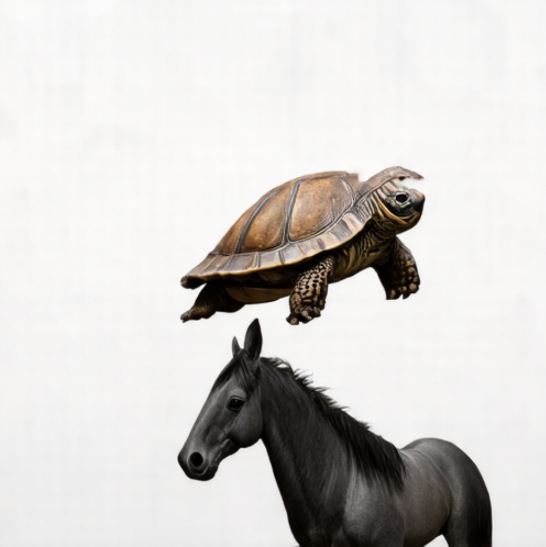} & \qi{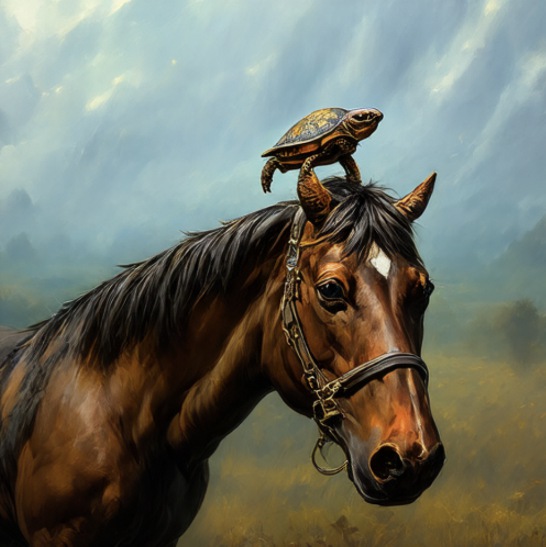} & \qi{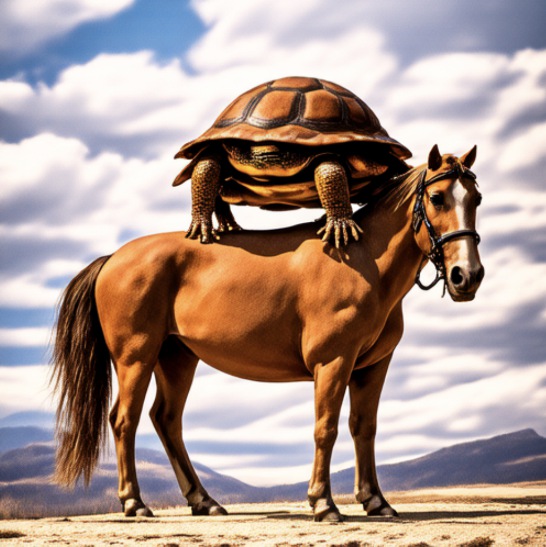} \\ \noalign{\vskip 2pt}
\qp{a photo of four boats} & \qi{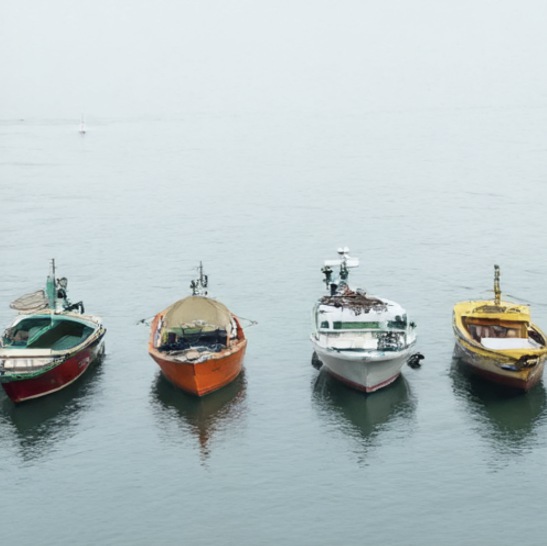} & \qi{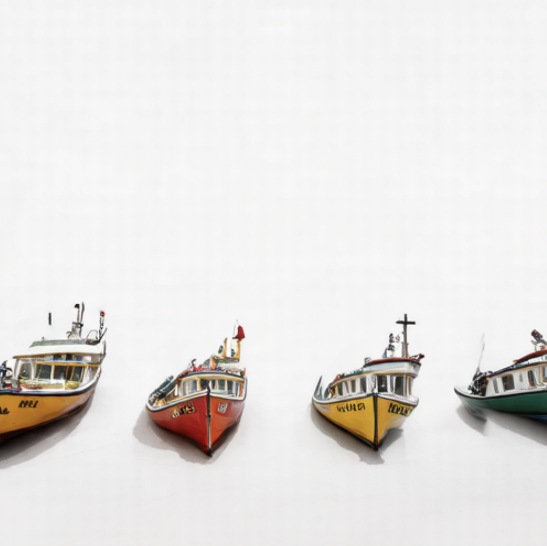} & \qi{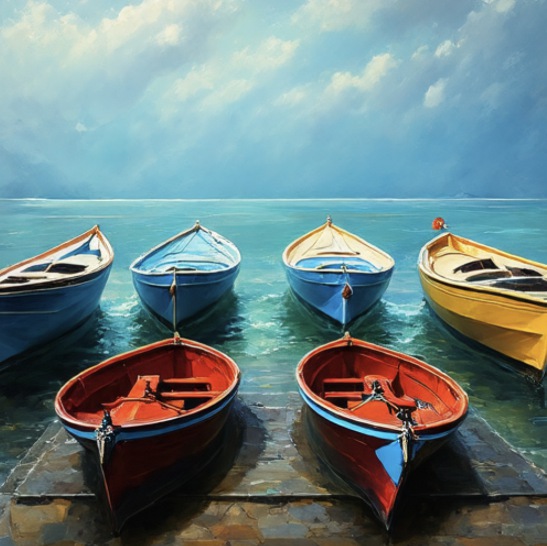} & \qi{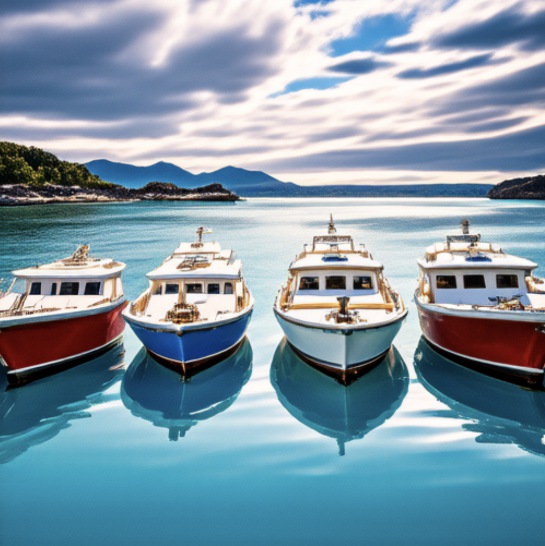} \\
\end{tabular}}

 \caption{\textbf{Qualitative comparisons of Flow-GRPO variants on \new{GenEval and GenEval2 prompts}.} Our model achieves a superior balance between visual quality and compositional accuracy.
    }
  \label{fig:qualitative}
\vspace{-12pt}
\end{figure*}

\Cref{tab:t2i-compbench} breaks down T2I-CompBench++ performance by
category. DeFlowCritic achieves the best scores on color, texture, and
3D-spatial relations, and remains close to Flow-GRPO on numeracy,
2D-spatial, and non-spatial categories. These gains are consistent with our
reward data construction, where negatives are created by controlled
perturbations of object attributes, bindings, and spatial relations.

\paragraph{Qualitative results.}
\Cref{fig:qualitative} compares generations from Flow-GRPO, Fast Flow-GRPO, Fast Flow-GRPO (multi-reward), and DeFlowCritic. 
Flow-GRPO often captures the requested composition but can produce less realistic images such as floating objects. 
Fast Flow-GRPO improves efficiency but  further degrades lighting, texture, and overall visual quality. 
Multi-reward Fast Flow-GRPO improves visual quality by incorporating aesthetic rewards, but often sacrifices compositional accuracy. 
In contrast, DeFlowCritic better preserves prompt-relevant details, while producing cleaner and more visually appealing images.

\begin{wraptable}{r}{0.64\textwidth}
\vspace{-6pt}
\centering
\footnotesize
\setlength{\tabcolsep}{3pt}
\renewcommand{\arraystretch}{1.08}
\caption{Comparison with latent reward models. Reward data is the training set of each reward model: Pick-a-Pic (LPO, SLRM) or HPDv3 (DINA-LRM). Rows marked $^\dagger$ are measured by us.}
\label{tab:latent_rewards}
\begin{tabular}{lllccc}
\toprule
Method & Backbone & Reward data & GenEval-1 & HPSv2.1 & HPSv3 \\
\midrule
SD3-M & SD3-M & -- & 67.29 & 28.74 & -- \\
LPO~\citep{zhang2025latentreward} & SD3-M & 0.17M pairs & 68.63 & 30.44 & -- \\
LPO, from~\citep{slrm2025consistent} & SDXL & 0.17M pairs & 59.85 & 31.89 & 9.10 \\
\midrule
SD3.5-M & SD3.5-M & -- & 55.34 & 27.15 & 8.29 \\
SLRM~\citep{slrm2025consistent} & SD3.5-M & 0.58M pairs & 68.93 & 32.01 & 9.89 \\
DINA-LRM$^\dagger$~\citep{liu2026dinalrm} & SD3.5-M & 0.8M pairs & 68.15 & 32.41 & 9.69 \\
\rowcolor{ours} DeFlowCritic$^\dagger$ & SD3.5-M & 5K images & 90.65 & 30.71 & 9.62 \\
\bottomrule
\end{tabular}
\vspace{-4pt}
\end{wraptable}

\paragraph{Comparison with latent reward models.}
We next compare DeFlowCritic with existing latent reward models by evaluating the policies obtained by finetuning/RL training with each of them. LPO~\citep{zhang2025latentreward}, SLRM~\citep{slrm2025consistent}, and DINA-LRM~\citep{liu2026dinalrm} learn from $0.17$M to $0.8$M human-preference pairs, while DeFlowCritic learns from only $5{,}007$ curated images with rule-graded caption perturbations. \Cref{tab:latent_rewards} reports compositional accuracy on GenEval-1 and preference scores (HPSv2.1 and HPSv3) on Pick-a-Pic prompts. DeFlowCritic reaches $90.65$ on GenEval-1, about $22$ points above the prior methods. On Pick-a-Pic, the home domain of preference rewards, it stays comparable on HPSv3 ($9.62$ vs.\ $9.69$ and $9.89$), while DINA-LRM and SLRM lead on HPSv2.1. DeFlowCritic thus reaches much higher compositional accuracy, with comparable visual quality while using substantially less reward data and no human-preference annotation.

\begin{wraptable}{r}{0.55\textwidth}
\vspace{-8pt}
\centering
\footnotesize
\setlength{\tabcolsep}{3pt}
\renewcommand{\arraystretch}{1.08}
\caption{Inference-time scaling on GenEval2. Calls are denoiser passes per image, on one H100.}
\label{tab:inference_ge2}
\begin{tabular}{lcccc}
\toprule
Method & Calls & sec/im & GenEval2 & HPSv3 \\
\midrule
SD3.5-M (base) & 80 & 1.4 & 18.10 & 6.66 \\
\;+ CLIPScore BO-8 & 640 & 7.3 & 21.95 & 6.58 \\
\midrule
\rowcolor{ours} \;+ DeFlowCritic BO-8 (ours) & 444 & 5.1 & 24.09 & 7.32 \\
\bottomrule
\end{tabular}
\vspace{-4pt}
\end{wraptable}

\paragraph{Inference-time scaling.}
We further evaluate the frozen critic as a test-time selection signal on GenEval2, using best-of-$K$ selection at $t^\star=602$ (\Cref{sec:stage2}).
\Cref{tab:inference_ge2} compares this with best-of-$K$ selection by CLIPScore, which must denoise and decode all $K$ candidates before it can rank them.
With $K=8$, DeFlowCritic reaches $24.09$ versus $21.95$ for CLIPScore, at $444$ instead of $640$ model calls and with higher HPSv3.

\subsection{Ablation Studies}
\label{sec:ablation}

We ablate the reward model $r_\phi$ design choices: \new{the backbone, the trainable components, the training objective, and the scoring timestep}. We report ranking accuracy (\textbf{$\mathrm{acc}_{\mathrm{rank}}$}), the percentage of evaluation rows
where the positive prompt is scored above all negatives paired with the same
reference image. \new{We also report pairwise accuracy ($\mathrm{acc}_{\mathrm{pair}}$), the percentage of prompt pairs for the same image with different target scores that the critic puts in the right order. Both are measured on our held-out test split (\Cref{sec:reward-data}): $240$ reference images that the critic never sees in training, with $1{,}422$ negatives built by the same rules as in the training set.}

\begin{table}[!h]
\centering
\small
\setlength{\tabcolsep}{6pt}
\renewcommand{\arraystretch}{1.15}
\caption{Ranking accuracy ($\mathrm{acc}_{\mathrm{rank}}$) (\%) of reward models on noisy intermediate latents.}
\label{tab:reward_vs_vlm}
\resizebox{\textwidth}{!}{%
\begin{tabular}{lcc|cccc}
\toprule
Method & Reward input & Fine-tuned & $t{=}465$ & $t{=}602$ & $t{=}707$ & $t{=}780$ \\
\midrule
Qwen-3-VL (zero-shot)              & Decoded image & $\times$     & 58.41 & 22.66 & 12.73 & 5.23  \\
Qwen-3-VL (fine-tuned, noisy images) & Decoded image & \checkmark   & 92.16 & 85.07 & 77.64 & 67.22 \\
\rowcolor{ours} DeFlowCritic & Latent & \checkmark & 97.08 & 96.67 & 95.83 & 91.03 \\
\bottomrule
\end{tabular}
\vspace{-12pt}
}
\end{table}

\paragraph{Backbone choice: SD3.5 versus VLM.}
A natural alternative to DeFlowCritic is to decode noisy latents into noisy
images and score them with a vision-language model. \Cref{tab:reward_vs_vlm}
compares zero-shot Qwen-3-VL~\cite{Qwen3-VL}, Qwen-3-VL fine-tuned on the same reward
dataset, and our SD3.5-based critic across four timesteps. Zero-shot
Qwen-3-VL rapidly fails as noise increases, dropping from $58.41\%$ at
$t=465$ to $5.23\%$ at $t=780$. Fine-tuning improves robustness, but still
falls to $67.22\%$ at high noise. In contrast, DeFlowCritic remains accurate
across the trajectory, achieving $97.08\%$, $96.67\%$, $95.83\%$, and
$91.03\%$. This supports the use of the generator backbone as the reward
representation: its features are pretrained on the same noisy latent space
visited during RL, whereas VLM features remain mismatched to intermediate
denoising states.

\begin{wraptable}{r}{0.6\textwidth}
\vspace{-6pt}
\centering
\footnotesize
\setlength{\tabcolsep}{3pt}
\renewcommand{\arraystretch}{1.05}
\caption{\new{LoRA and query tokens at $t=465$, with downstream GRPO on GenEval2.}}
\label{tab:critic_ablation}
\begin{tabular}{lcccc}
\toprule
Trainable parts & Params & $\mathrm{acc}_{\mathrm{rank}}$ & GenEval2 & HPSv3 \\
\midrule
Head only                    & 9.45M & 29.17 & 3.33  & $-$2.46 \\
Queries + head               & 9.48M & 27.50 & 2.38  & $-$9.89 \\
LoRA + head                  & 51.3M & 95.00 & 21.58 & 7.62 \\
\rowcolor{ours} LoRA + queries + head & 51.4M & 97.08 & 21.44 & 8.07 \\
\bottomrule
\end{tabular}
\vspace{-3pt}
\end{wraptable}

\paragraph{LoRA and query tokens.}
\Cref{tab:critic_ablation} isolates the trainable components. LoRA is the primary contributor: without backbone adaptation, the query tokens and reward head alone reach less than $30\%$ ranking accuracy and the downstream policy collapses. A critic that cannot rank also cannot guide training: without LoRA, the policy falls far below the base model on GenEval2 ($3.33$ and $2.38$, against $18.10$).

\begin{wraptable}{r}{0.47\textwidth}
\vspace{-6pt}
\centering
\small
\setlength{\tabcolsep}{4pt}
\renewcommand{\arraystretch}{1.08}
\caption{Loss ablation at $t=465$ on the test set. All variants use the same training step.}
\label{tab:reward_loss_ablation}
\begin{tabular}{lcc}
\toprule
Loss & $\mathrm{acc}_{\mathrm{rank}}$ & $\mathrm{acc}_{\mathrm{pair}}$ \\
\midrule
Binary BCE & 95.85 & 73.71 \\
Soft BCE only & 94.17 & 98.93 \\
Ranking only & 93.33 & 99.06 \\
Soft BCE + ranking & 95.42 & 99.06 \\
\rowcolor{ours} Soft BCE + ranking + FM & 97.08 & 99.05 \\
\bottomrule
\end{tabular}
\vspace{-4pt}
\end{wraptable}

\paragraph{Training objective.}
We next ablate the critic objective while keeping the architecture fixed. \Cref{tab:reward_loss_ablation} compares combinations of the training objectives, including binary BCE, which collapses all negatives to label $0$; soft BCE, which preserves the absolute continuous compositional scores; and the graded ranking loss which preserves the relative ranks. Binary BCE achieves high ranking accuracy, but much lower pairwise accuracy, indicating that binary labels discard severity information among negative prompts. Soft BCE improves pairwise accuracy by supervising the continuous score scale. Combining soft BCE, the graded ranking loss, and the flow-matching term gives both strong ranking and pairwise accuracy. The three losses therefore provide complementary supervision, and the full objective keeps all of them. 

\begin{wraptable}{r}{0.5\textwidth}
\vspace{-6pt}
\centering
\small
\setlength{\tabcolsep}{4pt}
\renewcommand{\arraystretch}{1.08}
\caption{ODE lookahead and the choice of $t^\star$.}
\label{tab:anchor}
\begin{tabular}{lccc}
\toprule
Noisy steps & $t^\star$ & GenEval2 & Rel. time \\
\midrule
No lookahead & -- & 20.41 & 1.00$\times$ \\
Lookahead & 278 & 22.06 & 1.72$\times$ \\
\rowcolor{ours} Lookahead & 602 & 21.44 & 1.26$\times$ \\
\bottomrule
\end{tabular}
\vspace{-4pt}
\end{wraptable}

\paragraph{Scoring timestep and lookahead.}
\Cref{tab:anchor} ablates the use of lookahead and the choice of $t^\star$. Without lookahead, the critic must evaluate highly noisy latents, which weakens the compositional supervision and reduces GenEval2 performance. Propagating farther to $t^\star=278$ slightly improves GenEval2, but introduces more computational effort. We therefore choose $t^\star=602$ empirically, as a compute-performance trade-off.

\begin{wraptable}[9]{r}{0.52\textwidth}
\vspace{-6pt}
\centering
\small
\setlength{\tabcolsep}{4pt}
\renewcommand{\arraystretch}{1.08}
\caption{Critic scores at $t=465$ on images that are all compositionally correct. $200$ images per pool.}
\label{tab:quality_source}
\begin{tabular}{lccc}
\toprule
Image pool & Correct & Critic & HPSv3 \\
\midrule
GPT-Image-1.5 (reference) & 100\% & 0.97 & 10.93 \\
\rowcolor{ours} Our GRPO policy & 100\% & 0.96 & 10.71 \\
SD3.5-M (base model) & 100\% & 0.82 & 9.66 \\
\bottomrule
\end{tabular}
\vspace{-8pt}
\end{wraptable}

\paragraph{Critic's preference on image quality.}
DeFlowCritic receives no explicit supervision for image quality.
Each negative keeps the reference image and corrupts only its prompt, so the critic never sees one image labeled as worse than another. To understand what the critic rewards beyond composition, we take its held-out test prompts and collect one image per prompt from three sources: the GPT-Image-1.5 reference images, SD3.5-M, the base model of both the critic and the policy; and our GRPO policy. A VLM judge then keeps only the images it rates as fully correct, leaving $200$ images per source that all match their prompts and differ only in how they look. We encode each image, noise it to $t=465$, and score it against its prompt with the frozen critic (\Cref{tab:quality_source}). Although every image is correct, the critic does not score them equally: the references receive $0.97$ and our policy $0.96$, while SD3.5-M receives only $0.82$. The critic therefore favors images from a distribution close to its references, even over images from its own base model. GRPO moves the policy toward this distribution, and HPSv3 rises from $9.66$ to $10.71$ along the way.

\needspace{10\baselineskip}
\subsection{Generalization Across T2I Models}
\label{sec:generators}

\begin{wraptable}[8]{r}{0.45\textwidth}
\vspace{-22pt}
\centering
\small
\setlength{\tabcolsep}{4pt}
\renewcommand{\arraystretch}{1.08}
\caption{\new{DeFlowCritic GRPO improves different T2I flow matching models.}}
\label{tab:other_generators}
\begin{tabular}{lccc}
\toprule
Generator & GenEval2 & HPSv3 \\
\midrule
SD3.5-L & 19.14 & 3.93  \\
\rowcolor{ours} + GRPO, 440 steps & 25.83 & 8.29  \\
\midrule
FLUX.1-dev & 21.40 & 7.89  \\
\rowcolor{ours} + GRPO, 240 steps & 28.49 & 9.02  \\
\bottomrule
\end{tabular}
\end{wraptable}

\begin{figure}[t]
\centering
\scriptsize
\setlength{\tabcolsep}{1pt}
\renewcommand{\arraystretch}{0.5}
\def\fq#1{\raisebox{-.5\height}{\includegraphics[width=0.15\textwidth]{camera_ready/figures/flux_qual/#1}}}
\def\fp#1{\parbox[t]{0.15\textwidth}{\centering\emph{#1}}}
\begin{tabular}{@{}c@{\hspace{2pt}}cccccc@{}}
\rotatebox[origin=c]{90}{FLUX.1-dev} & \fq{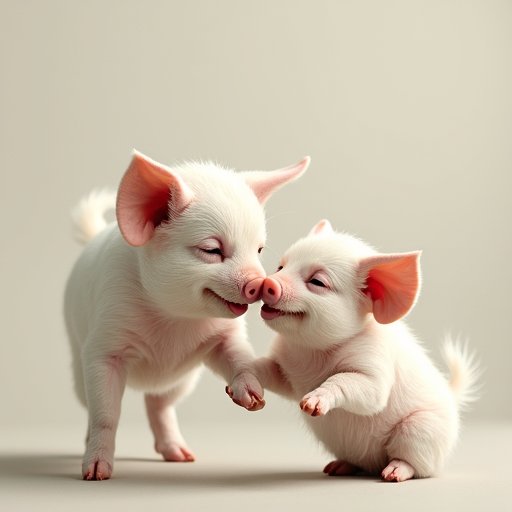} & \fq{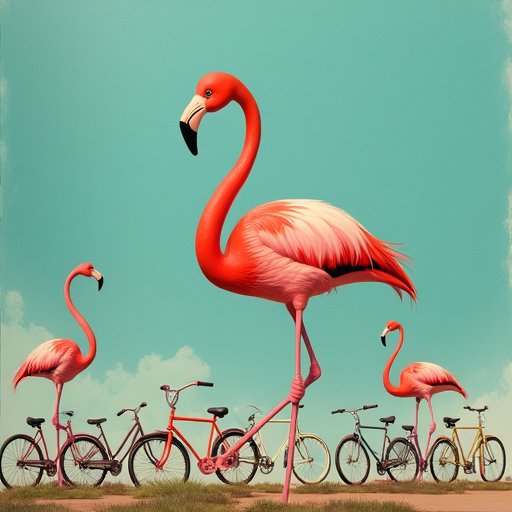} & \fq{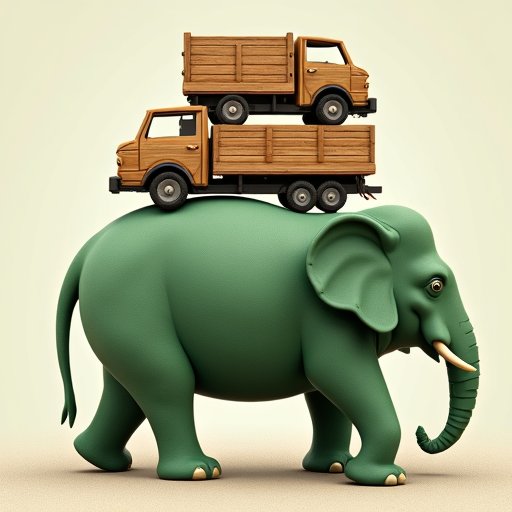} & \fq{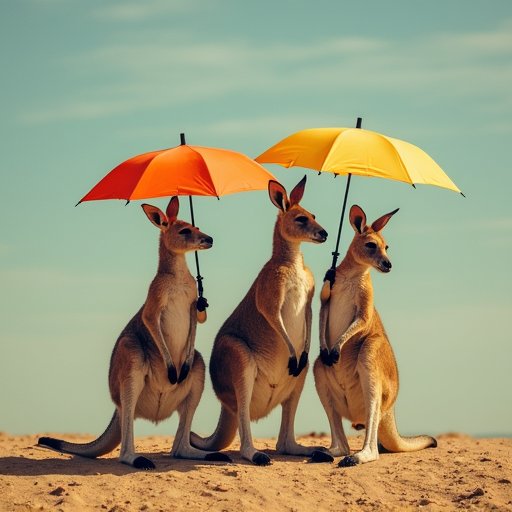} & \fq{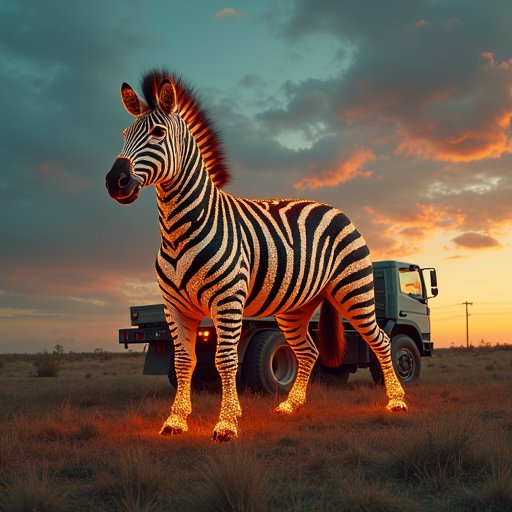} & \fq{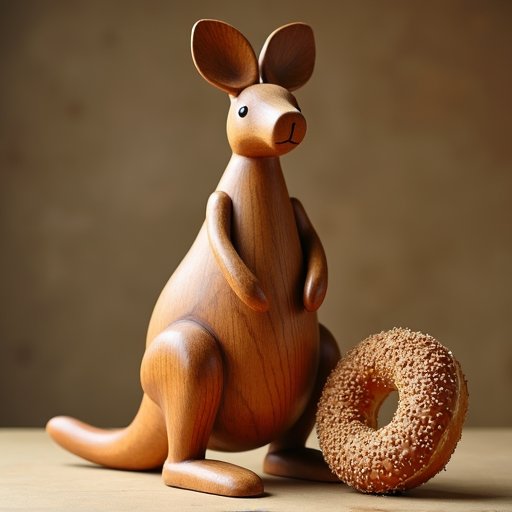} \\ \noalign{\vskip 2pt}
\rotatebox[origin=c]{90}{+ GRPO (ours)} & \fq{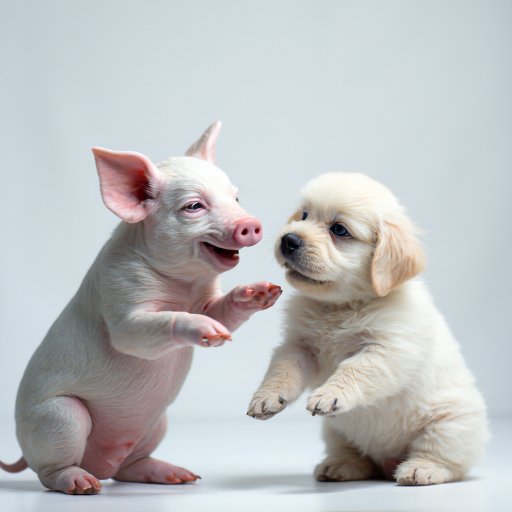} & \fq{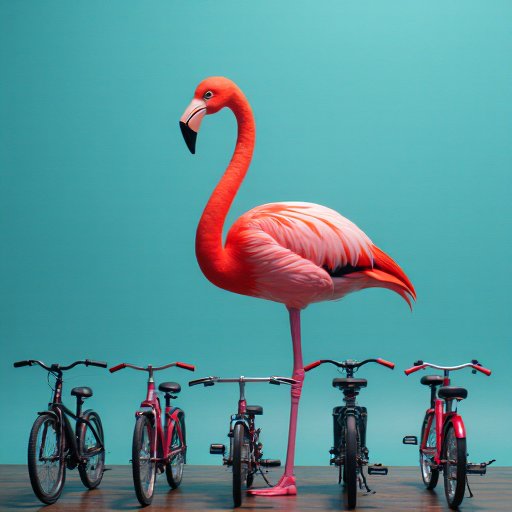} & \fq{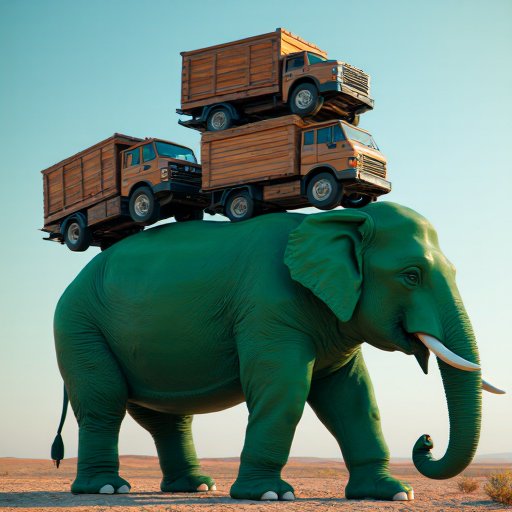} & \fq{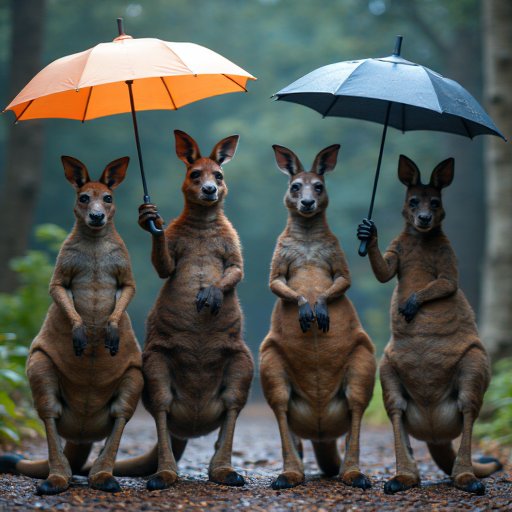} & \fq{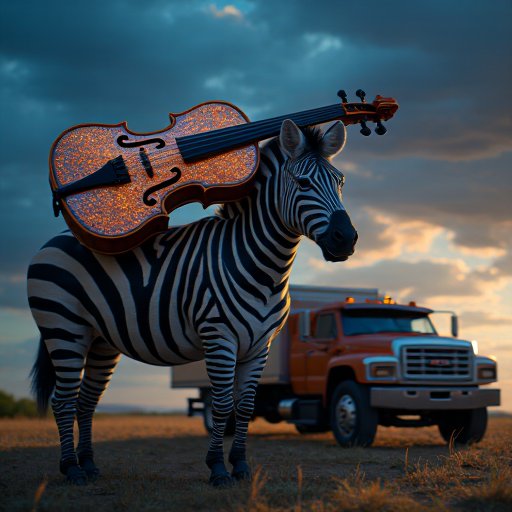} & \fq{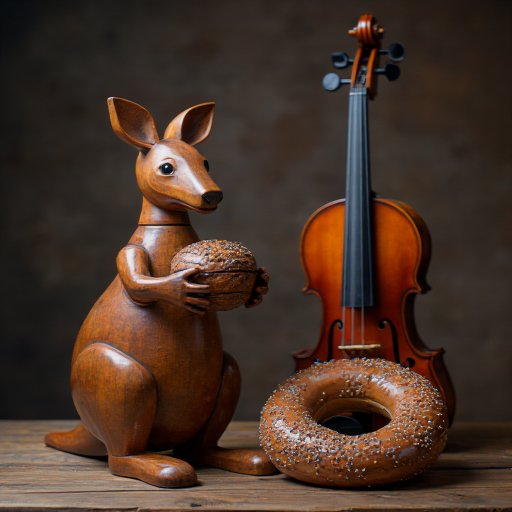} \\ \noalign{\vskip 4pt}
 & \fp{a white pig playing with a white dog} & \fp{a flamingo and five bicycles} & \fp{three wooden trucks on top of a green elephant} & \fp{four kangaroos under two umbrellas} & \fp{a sparkling violin on top of a zebra, and a truck} & \fp{a wooden kangaroo and a bagel and a violin} \\
\end{tabular}
\caption{\new{FLUX.1-dev before (top) and after (bottom) $240$ GRPO steps with DeFlowCritic, evaluated on GenEval2 prompts.}}
\label{fig:flux}
\vspace{-14pt}
\end{figure}

We evaluate DeFlowCritic across T2I models in two settings.
When generators share the same VAE latent space, the same critic can be reused without retraining. We demonstrate this by applying our SD3.5-Medium critic, built on a $2.5$B backbone, to guide an $8$B SD3.5-Large policy.
When a generator uses a different latent space, a new critic is required, but the same training approach remains applicable. For FLUX.1-dev~\citep{flux2024}, we train a FLUX-based critic using the same $5$K reference images and training objective. Training takes $837$ steps, approximately $5$ H100-hours (\Cref{app:critic-training}).
As shown in \Cref{tab:other_generators}, DeFlowCritic-guided GRPO improves GenEval2 scores from $19.14$ to $25.83$ on SD3.5-Large and from $21.40$ to $28.49$ on FLUX.1-dev, and increases HPSv3 as well.
\Cref{fig:flux} shows examples where base FLUX misses an object, a count, or a relation, and GRPO fixes the compositions while retaining the visual quality.

\FloatBarrier
\section{Conclusion}
\label{sec:conclusion}

We presented \textbf{DeFlowCritic}, a dense latent reward model for efficient compositional alignment of text-to-image flow-matching models. Instead of relying on sparse terminal rewards from decoded images, DeFlowCritic scores noisy intermediate latents directly using the generator's own representation space, enabling timestep-specific feedback during online GRPO training. Experiments on GenEval2 and T2I-CompBench++ show that the method preserves the compositional gains of Flow-GRPO, improves image-quality metrics such as HPSv3, and substantially reduces training cost; the same frozen critic also improves inference-time trajectory selection without additional training. These results suggest that generator-native latent critics are a promising direction for scalable, reward-efficient alignment of diffusion and flow-matching generators.

\paragraph{Limitations.}
\label{sec:limitaion}
Our experiments focus on compositional T2I; transfer to higher resolutions and
video or image-to-video generation remains open. The method also depends on the empirical finding that early-to-mid
denoising latents are effective scoring states, so other noise schedules may require adaptive selection of $t^\star$.
\paragraph{Broader impact.}
\label{sec:impact}
While stronger compositional control
can benefit creative and design applications, it may also increase misuse risks
for deceptive or harmful synthetic imagery; responsible deployment should pair
latent reward alignment with content filtering, dataset auditing, and
provenance mechanisms.

\begin{ack}
This work was granted access to the HPC resources of IDRIS under the allocation AD011014688R2 made by GENCI. It was funded in part by the French government under management of Agence Nationale de la Recherche as part of the ``France 2030'' program, reference ANR-23-IACL-0008 (PR{[}AI{]}RIE-PSAI project), and Paris \^{I}le-de-France R\'{e}gion in the frame of the DIM AI4IDF. Cordelia Schmid would like to acknowledge the support by the K\"{o}rber European Science Prize.
\end{ack}

{
\small
\bibliographystyle{plain}
\bibliography{camera_ready/references}
}

\clearpage
\appendix

\section{Scoring Rules}
\label{app:rules}

\paragraph{Prompt specifications.}
Each prompt is generated from a structured specification of its objects, attributes, counts, relation, and scene based on GenEval-1 compositions.
The four categories use different parts of this specification. A counting prompt fixes the number of instances of one object class, a color prompt gives one object a color, an attribute-binding prompt names two objects with one color each, and a position prompt places two objects in a spatial relation. Every prompt also carries a scene phrase, such as \emph{at a train station}.
During critic training, half of the batches use the prompts with their scene phrase and half without it, so the critic sees both object-specific short prompts and longer prompts that adds a background.

\paragraph{Negatives and their scores.}
A negative prompt applies one corruption to this specification and is created with the same template, so it differs from the positive only in the corrupted part.
The corruption type alone determines its score: the more of the prompt that remains correct for the reference image, the higher the score.
The labels are therefore fully rule-based, with no per-sample human or LLM judgment.
\Cref{tab:rules} lists every corruption type with a real example. Three principles set the scores:
\begin{itemize}
\item \textbf{Count errors are graded by their number.} A count that is off by one, two, or three instances scores $0.75$, $0.5$, or $0.25$.
\item \textbf{Attribute errors keeping the objects correct.} A wrong color, relation, or scene scores $0.5$, and $0.75$ when only one of two colors in attribute binding is wrong.
\item \textbf{Object errors are the most severe.} Replacing one of two objects scores $0.25$, and replacing every object scores $0$.
\end{itemize}

\begin{table}[h]
\centering
\footnotesize
\setlength{\tabcolsep}{4pt}
\renewcommand{\arraystretch}{1.1}
\caption{Corruption types with real examples from the dataset.}
\label{tab:rules}
\begin{tabular}{lllcr}
\toprule
Category & Corruption & Example negative & $y$ & Negatives \\
\midrule
\multirow{3}{*}{\makecell[l]{Counting\\\emph{Four trains at a train station}}}
 & Count off by one & Three trains at a train station & 0.75 & 1,624 \\
 & Count off by two & Two trains at a train station & 0.50 & 1,008 \\
 & Count off by three & One train at a train station & 0.25 & 392 \\
\midrule
\multirow{2}{*}{\makecell[l]{Color\\\emph{an orange oven in a modern kitchen}}}
 & Wrong color & a blue oven in a modern kitchen & 0.50 & 3,753 \\ \\
\midrule
\multirow{4}{*}{\makecell[l]{Attribute binding\\\emph{A yellow broccoli and a}\\\emph{red tennis racket}}}
 & One color wrong & A purple broccoli and a red tennis racket & 0.75 & 2,910 \\
 & Both colors wrong & A black broccoli and a brown tennis racket & 0.50 & 1,455 \\
 & One object & A brown giraffe and a red tennis racket & 0.25 & 2,910 \\
 & All objects & A pink baseball bat and a brown tv remote & 0.00 & 1,455 \\
\midrule
\multirow{3}{*}{\makecell[l]{Position\\\emph{A fork above a cow}}}
 & Wrong relation & A fork left of a cow & 0.50 & 3,879 \\
 & One object replaced & A snowboard above a cow & 0.25 & 2,586 \\
 & Both objects replaced & A hair drier above a potted plant & 0.00 & 1,293 \\
\midrule
\multirow{2}{*}{\makecell[l]{All categories\\\emph{Three trains at a train station}}}
 & Scene replaced & Three trains in a kitchen & 0.50 & 4,792 \\
 & Only object replaced & Three cats at a train station & 0.00 & 2,259 \\
\bottomrule
\end{tabular}

\end{table}

\paragraph{Dataset statistics.}
\Cref{tab:data_stats} breaks the dataset down by category. The $5{,}007$ training rows carry $30{,}316$ negatives, between $4$ and $7$ per image and $6.05$ on average. Counting and color rows have about five negatives each, and the two-object categories about seven, because they have more parts to corrupt. Every image has exactly one fully wrong negative with $y=0$, so each row spans the whole score range. The test split has $60$ rows per category, $240$ in total, with $1{,}422$ negatives built by the same rules. Its images are never used for training. 

\begin{table}[h]
\centering
\footnotesize
\setlength{\tabcolsep}{4pt}
\renewcommand{\arraystretch}{1.1}
\caption{Rows and negatives per category, with the number of training negatives at each score $y$.}
\label{tab:data_stats}
\begin{tabular}{lrrrrrrrrr}
\toprule
 & \multicolumn{7}{c}{Training} & \multicolumn{2}{c}{Test} \\
\cmidrule(lr){2-8} \cmidrule(lr){9-10}
Category & Rows & Negatives & Per image & $y{=}0.75$ & $y{=}0.5$ & $y{=}0.25$ & $y{=}0$ & Rows & Negatives \\
\midrule
Counting          & 1,008 & 5,018  & 4.98 & 1,624 & 1,994  & 392   & 1,008 & 60  & 297 \\
Color             & 1,251 & 6,164  & 4.93 & --    & 4,913  & --    & 1,251 & 60  & 295 \\
Attribute binding & 1,455 & 10,114 & 6.95 & 2,910 & 2,839  & 2,910 & 1,455 & 60  & 414 \\
Position          & 1,293 & 9,020  & 6.98 & --    & 5,141  & 2,586 & 1,293 & 60  & 416 \\
\midrule
Total             & 5,007 & 30,316 & 6.05 & 4,534 & 14,887 & 5,888 & 5,007 & 240 & 1,422 \\
\bottomrule
\end{tabular}
\end{table}

\FloatBarrier
\section{Unseen-Prompt Evaluation Set}
\label{app:ood}

Our critic is initialized from a large pretrained diffusion model with broad knowledge of visual concepts and their compositions, so LoRA fine-tuning mainly aligns the pretrained representations with compositional scoring rather than relearning or overwriting this knowledge.
Although our corruption rules use controlled semantic edits, they require fine-grained comparisons involving objects, attributes, counts, spatial relations, and scenes.
We therefore expect this compositional supervision to transfer to open-domain prompts and new perturbations.

To test this, we construct a new evaluation set from user-written prompts in Pick-a-Pic~\citep{kirstain2023pickapic}.
These prompts are free-form and include artistic styles, action verbs, surreal concepts, and world knowledge, none of which are present in our training prompts, for example \emph{A fisherman fishing on a no man's lake in 1000 years}, \emph{tshirt design with a zebra colorful rock and roll}, and \emph{transylvania castle on hilltop and dusk bats and scifi}.
For each prompt, an LLM extracts five semantic specifications (objects, actions, attributes, counts, and scenes) and generates negative captions by modifying one or more of them.
We then generate images for the positive prompts with GPT-Image-1.5 and manually filter the resulting pairs for correctness, which leaves $200$ positive prompts and $716$ negatives.
\Cref{tab:ood_set} shows an example.

\begin{table}[h]
\centering
\small
\setlength{\tabcolsep}{6pt}
\renewcommand{\arraystretch}{1.1}
\caption{An example from the unseen-prompt set.}
\label{tab:ood_set}
\begin{tabular}{cll}
\toprule
Label & Kind & Caption \\
\midrule
1.00 & Positive & Crazy monkey stealing food from gorilla \\
0.75 & Swap verb & Crazy monkey sharing food with gorilla \\
0.50 & Change count & Three crazy monkeys stealing food from gorilla \\
0.50 & Swap object & Crazy monkey stealing food from a bear \\
0.00 & Swap verb and objects & Calm parrot sharing food with a bear \\
\bottomrule
\end{tabular}
\end{table}

\Cref{tab:ood_results} reports the ranking accuracy on this set.
The FLUX critic achieves $91.6\%$ and the SD3.5-M critic $88.7\%$ at $t=465$, showing that compositional accuracy generalizes to unseen prompts and new perturbations.

\begin{table}[h]
\centering
\small
\setlength{\tabcolsep}{6pt}
\renewcommand{\arraystretch}{1.1}
\caption{Ranking accuracy ($\mathrm{acc}_{\mathrm{rank}}$, \%) on the unseen-prompt set.}
\label{tab:ood_results}
\begin{tabular}{lcc}
\toprule
$t$ & SD3.5-M critic & FLUX.1-dev critic \\
\midrule
465 & 88.7 & 91.6 \\
602 & 89.2 & 89.8 \\
707 & 89.2 & 91.0 \\
\bottomrule
\end{tabular}
\end{table}

\section{Critic Training}
\label{app:critic-training}

\paragraph{Recipe.}
Both critics follow the same recipe on the same $5{,}007$ training rows, and only the backbone changes.
The backbone stays frozen in bf16. We train LoRA adapters on its attention and feed-forward layers, the $16$ query tokens, and the reward head with AdamW and a cosine learning-rate schedule that decays to $0.1$ of its initial value.
Each micro-batch holds two reference images. Each image is paired with its positive prompt and all of its negatives, and all of these image–prompt pairs, about $14$, pass through the backbone together in one forward pass, sharing the image's latent and noise. One optimizer step accumulates gradients over nine micro-batches. For each micro-batch, we draw one timestep uniformly from $[50, 950]$ and noise all of its latents to it.
We set $\lambda_{\mathrm{rank}}=0.2$ with temperature $\tau=0.15$, and $\lambda_{\mathrm{FM}}=1$.
FLUX.1-dev is guidance-distilled, so its critic receives a constant guidance embedding of $3.5$. It also uses a higher LoRA learning rate, $10^{-4}$, which converged fast.
\Cref{tab:critic_training} summarizes both settings.

\paragraph{Stopping and cost.}
We use early stopping on the held-out split, and stop training once ranking accuracy plateaus.
The SD3.5-M critic reaches $97\%$ after $5{,}301$ steps, about $8$ hours on one H100, and the FLUX critic reaches $95\%$ after $837$ steps, about $5.2$ hours.
Although the FLUX critic takes longer per step, it converges in far fewer steps, which we attribute to its higher LoRA learning rate and guidance distillation.

\begin{table}[h]
\centering
\small
\setlength{\tabcolsep}{6pt}
\renewcommand{\arraystretch}{1.1}
\caption{Critic training settings and cost on one H100.}
\label{tab:critic_training}
\begin{tabular}{lcc}
\toprule
 & SD3.5-M critic & FLUX.1-dev critic \\
\midrule
Backbone (frozen) & SD3.5-M, 2.5B & FLUX.1-dev, 12B \\
LoRA & rank $32$, $\alpha=64$ & rank $32$, $\alpha=64$ \\
Query tokens & $16$ & $16$ \\
Learning rate (head / LoRA) & $10^{-4}$ / $3{\times}10^{-5}$ & $10^{-4}$ / $10^{-4}$ \\
Images per optimizer step & $2 \times 9$ & $2 \times 9$ \\
Training timesteps & $[50, 950]$ & $[50, 950]$ \\
Guidance embedding & -- & $3.5$ \\
\midrule
Time per optimizer step & 5.3 s & 22.4 s \\
Optimizer steps & 5,301 & 837 \\
Training time & $\sim$8 h & $\sim$5.2 h \\
\bottomrule
\end{tabular}
\end{table}

\newpage
\section*{NeurIPS Paper Checklist}

\begin{enumerate}

\item {\bf Claims}
    \item[] Question: Do the main claims made in the abstract and introduction accurately reflect the paper's contributions and scope?
    \item[] Answer: \answerYes{}
    \item[] Justification: The abstract and introduction state the paper's main contributions: DeFlowCritic, a dense latent reward model for flow-matching T2I alignment; its use for online GRPO training and inference-time trajectory selection; and empirical evaluation on GenEval2 and T2I-CompBench++\new{, with transfer to SD3.5-Large and FLUX.1-dev}. These claims are supported by the method and experiments in Sections~\ref{sec:method}\new{,}~\ref{sec:sota}\new{, and~\ref{sec:generators}}.

\item {\bf Limitations}
    \item[] Question: Does the paper discuss the limitations of the work performed by the authors?
    \item[] Answer: \answerYes{}
    \item[] Justification: The paper discusses limitations and broader impact in Section~\ref{sec:impact}, including transfer to higher resolutions, video or image-to-video generation, and the empirical choice of the trusted scoring timestep.

\item {\bf Theory assumptions and proofs}
    \item[] Question: For each theoretical result, does the paper provide the full set of assumptions and a complete (and correct) proof?
    \item[] Answer: \answerNA{}
    \item[] Justification: The paper does not present theoretical results or formal proofs; it proposes an empirical method for dense latent reward modeling and online RL alignment.

\item {\bf Experimental result reproducibility}
    \item[] Question: Does the paper fully disclose all the information needed to reproduce the main experimental results of the paper to the extent that it affects the main claims and/or conclusions of the paper, regardless of whether the code and data are provided or not?
    \item[] Answer: \answerYes{}
    \item[] Justification: The paper provides the information needed to reproduce the main experiments, including reward-model dataset construction and split, model architecture, training objective, key hyperparameters, GRPO training setup, evaluation benchmarks, and compute cost in Sections~\ref{sec:method} and~\ref{sec:impl} and Appendices~\ref{app:rules} and~\ref{app:critic-training}. Code, data, and model checkpoints will also be released \new{on the project page}.

\item {\bf Open access to data and code}
    \item[] Question: Does the paper provide open access to the data and code, with sufficient instructions to faithfully reproduce the main experimental results, as described in supplemental material?
    \item[] Answer: \answerNo{}
    \item[] Justification: \new{We will release the code, the reward-modeling dataset, and the model checkpoints on the project page (\url{https://zeeshank95.github.io/deflowcritic}); they are not yet public at the time of this version.} The paper nevertheless provides the main implementation details needed to understand and reproduce the experiments.

\item {\bf Experimental setting/details}
    \item[] Question: Does the paper specify all the training and test details, e.g., data splits, hyperparameters, how they were chosen, type of optimizer, necessary to understand the results?
    \item[] Answer: \answerYes{}
    \item[] Justification: The paper reports the main training and evaluation details, including data splits, model architecture, loss terms, hyperparameters, optimizer, timestep ranges, GRPO settings, and benchmark protocols in Sections~\ref{sec:reward-data} and~\ref{sec:impl}\new{, with the scoring rules and critic training details in Appendices~\ref{app:rules} and~\ref{app:critic-training}}. Additional low-level implementation details will be included with the released code.

\item {\bf Experiment statistical significance}
    \item[] Question: Does the paper report error bars suitably and correctly defined or other appropriate information about the statistical significance of the experiments?
    \item[] Answer: \answerNo{}
    \item[] Justification: We do not report error bars or confidence intervals because the main online RL experiments are computationally expensive. We instead report comparisons across multiple held-out benchmarks and ablations. Repeated runs and statistical uncertainty estimates are left for future work.

\item {\bf Experiments compute resources}
    \item[] Question: For each experiment, does the paper provide sufficient information on the computer resources, type of compute workers, memory, time of execution, needed to reproduce the experiments?
    \item[] Answer: \answerYes{}
    \item[] Justification: Section~\ref{sec:impl} reports the compute resources and runtime for the main experiments, including H100 GPU usage, DeFlowCritic training time, and GRPO training time. \new{Appendix~\ref{app:critic-training} gives the training cost of each critic, and Table~\ref{tab:inference_ge2} the inference cost of best-of-$K$ selection.}

\item {\bf Code of ethics}
    \item[] Question: Does the research conducted in the paper conform, in every respect, with the NeurIPS Code of Ethics \url{https://neurips.cc/public/EthicsGuidelines}?
    \item[] Answer: \answerYes{}
    \item[] Justification: The research conforms to the NeurIPS Code of Ethics. The work uses generated and benchmark data for text-to-image alignment research, and we discuss potential misuse risks and responsible deployment considerations in the limitations and broader impact discussion.

\item {\bf Broader impacts}
    \item[] Question: Does the paper discuss both potential positive societal impacts and negative societal impacts of the work performed?
    \item[] Answer: \answerYes{}
    \item[] Justification: The limitations and broader impact discussion in Section~\ref{sec:impact} covers both positive impacts, such as improved prompt faithfulness for creative and design applications, and negative impacts, including potential misuse for deceptive or harmful synthetic imagery. We also discuss responsible deployment measures such as content filtering, dataset auditing, and provenance mechanisms.

\item {\bf Safeguards}
    \item[] Question: Does the paper describe safeguards that have been put in place for responsible release of data or models that have a high risk for misuse, e.g., pre-trained language models, image generators, or scraped datasets?
    \item[] Answer: \answerNo{}
    \item[] Justification: We discuss responsible deployment considerations, including content filtering, dataset auditing, and provenance mechanisms, but the current paper does not specify concrete release safeguards for the model or dataset. We will include usage guidelines and release documentation with the public artifacts.

\item {\bf Licenses for existing assets}
    \item[] Question: Are the creators or original owners of assets, e.g., code, data, models, used in the paper, properly credited and are the license and terms of use explicitly mentioned and properly respected?
    \item[] Answer: \answerYes{}
    \item[] Justification: We cite the existing models, benchmarks, and evaluation assets used in the paper, including SD3.5-Medium, \new{SD3.5-Large, FLUX.1-dev,} Flow-GRPO, \new{GenEval,} GenEval2, T2I-CompBench++, \new{Pick-a-Pic,} LAION Aesthetic, \new{HPSv2.1,} HPSv3, Qwen-3-VL\new{, and the DINA-LRM checkpoint}. We will include license and terms-of-use information for these assets in the supplementary material or public release, and will ensure that any released code, data, and models respect the corresponding licenses.

\item {\bf New assets}
    \item[] Question: Are new assets introduced in the paper well documented and is the documentation provided alongside the assets?
    \item[] Answer: \answerYes{}
    \item[] Justification: The paper introduces a new reward-modeling dataset and the DeFlowCritic model. Dataset construction, filtering, splits, label semantics, model architecture, and training objectives are documented in the paper, and additional documentation will be provided with the released assets.

\item {\bf Crowdsourcing and research with human subjects}
    \item[] Question: For crowdsourcing experiments and research with human subjects, does the paper include the full text of instructions given to participants and screenshots, if applicable, as well as details about compensation, if any?
    \item[] Answer: \answerNA{}
    \item[] Justification: The paper does not involve crowdsourcing experiments or human-subject research. The reward-modeling dataset is generated and manually filtered by the authors, without collecting responses from external participants.

\item {\bf Institutional review board (IRB) approvals or equivalent for research with human subjects}
    \item[] Question: Does the paper describe potential risks incurred by study participants, whether such risks were disclosed to the subjects, and whether Institutional Review Board (IRB) approvals, or an equivalent approval/review based on the requirements of your country or institution, were obtained?
    \item[] Answer: \answerNA{}
    \item[] Justification: The paper does not involve human-subject experiments or external study participants, so IRB approval or equivalent review is not applicable.

\item {\bf Declaration of LLM usage}
    \item[] Question: Does the paper describe the usage of LLMs if it is an important, original, or non-standard component of the core methods in this research? Note that if the LLM is used only for writing, editing, or formatting purposes and does \emph{not} impact the core methodology, scientific rigor, or originality of the research, declaration is not required.
    \item[] Answer: \answerYes{}
    \item[] Justification: We use GPT-Image-1.5 to generate reference images for the reward-modeling dataset, as described in Section~\ref{sec:reward-data}. This usage is part of the data construction pipeline and is disclosed in the paper. \new{We also use an LLM to write the negative captions of the unseen-prompt evaluation set (Appendix~\ref{app:ood}).}

\end{enumerate}

\end{document}